\pdfoutput=1
\documentclass[11pt]{article}

\usepackage[]{EMNLP2022}

\usepackage{times}
\usepackage{latexsym}

\usepackage{graphicx}
\usepackage{enumitem}
\usepackage{caption}
\usepackage{multirow}
\usepackage{array}
\usepackage{subcaption}
\usepackage{dirtytalk}
\usepackage{xspace}
\usepackage{booktabs}
\usepackage{amsmath}
\usepackage{amssymb}
\usepackage{balance}

\newcommand{\ie}{\emph{i.e.,}\xspace}
\newcommand{\eg}{\emph{e.g.,}\xspace}

\newcommand{\mname}{SMiLe-OIE\xspace}
\newcommand{\jj}[1]{{\textcolor{red}{#1}}}

\usepackage[T1]{fontenc}
\usepackage[utf8]{inputenc}

\usepackage{microtype}

\usepackage{inconsolata}
\usepackage[normalem]{ulem}

\title{Syntactic Multi-view Learning for Open Information Extraction}

\author{Kuicai Dong$^{1,2}$, Aixin Sun$^1$, Jung-Jae Kim$^2$, Xiaoli Li$^{1,2,3}$ \\
\fontsize{11pt}{12pt}\selectfont
$^1$ School of Computer Science and Engineering, Nanyang Technological University, Singapore\\
\fontsize{11pt}{12pt}\selectfont
\texttt{kuicai001@e.ntu.edu.sg, axsun@ntu.edu.sg}\\
\fontsize{11pt}{12pt}\selectfont
$^2$ Institute for Infocomm Research, A*STAR, Singapore\\
\fontsize{11pt}{12pt}\selectfont
$^3$ A*STAR Centre for Frontier AI Research, Singapore\\
\fontsize{11pt}{12pt}\selectfont
\texttt{\{jjkim, xlli\}@i2r.a-star.edu.sg}}

\begin{document}
\maketitle
\begin{abstract}

Open Information Extraction (OpenIE) aims to extract relational tuples from open-domain sentences. 
Traditional rule-based or statistical models have been developed based on syntactic structures of sentences, identified by syntactic parsers. However, previous neural OpenIE models under-explore the useful syntactic information.
In this paper, we model both \textit{constituency} and \textit{dependency} trees into word-level graphs, and enable neural OpenIE to learn from the syntactic structures. 
To better fuse heterogeneous information from both graphs, we adopt multi-view learning to capture multiple relationships from them. Finally, the finetuned constituency and dependency representations are aggregated with sentential semantic representations for tuple generation. 
Experiments show that both constituency and dependency information, and the multi-view learning are effective. Our model is publicly available.\footnote{https://github.com/daviddongkc/smile\_oie}

\end{abstract}

%======================================================================
\section{Introduction}\label{sec:intro}
%======================================================================

Open Information Extraction (OpenIE) aims to generate structured tuples from unstructured open-domain text~\cite{yates2007textrunner}.
The extracted tuples are in the form of $\langle Subject, Relation, Object \rangle$ for binary relation, and $\langle ARG_0, Relation, ARG_1, \dots, ARG_n \rangle$ for $n$-ary relation.
It has been a critical NLP task as it is domain-independent  and does not rely on pre-defined ontology schema. The structured relational tuples are beneficial to many  downstream tasks such as question answering~\cite{khot2017answering}, knowledge base population~\cite{martinez2018openie, gashteovski2020aligning} and word embedding generation~\cite{stanovsky2015open}.

\begin{figure}[t]
\centering
\begin{subfigure}[b]{1\linewidth}
   \centering
   \includegraphics[width=1\linewidth]{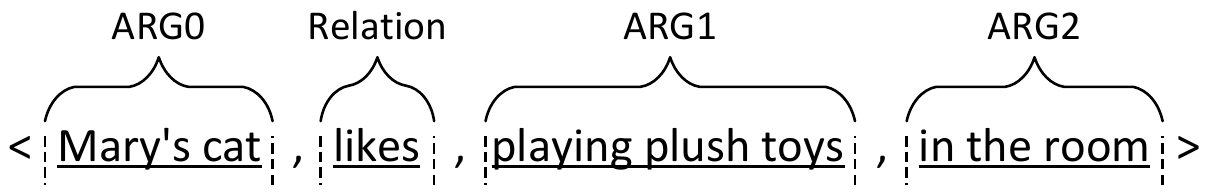}
   \caption{$n$-ary OpenIE tuple.}
   \label{fig:tuple_extraction}
\end{subfigure}
\begin{subfigure}[b]{1\linewidth}
   \centering
   \includegraphics[width=1\linewidth]{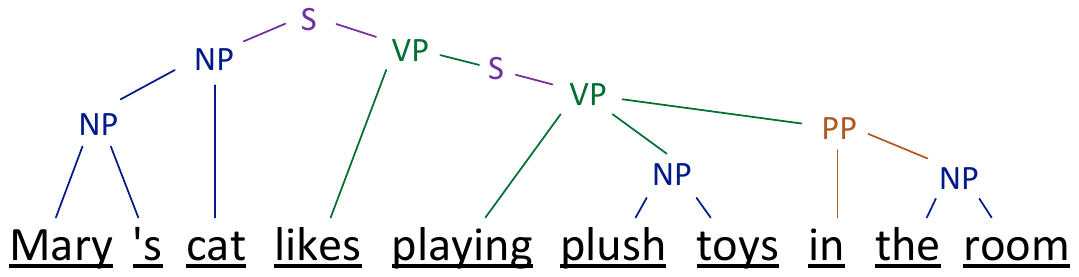}
   \caption{Constituency tree (results from CoreNLP).}
   \label{fig:const_tree}
\end{subfigure}
\begin{subfigure}[b]{1\linewidth}
   \centering
   \includegraphics[width=1\linewidth]{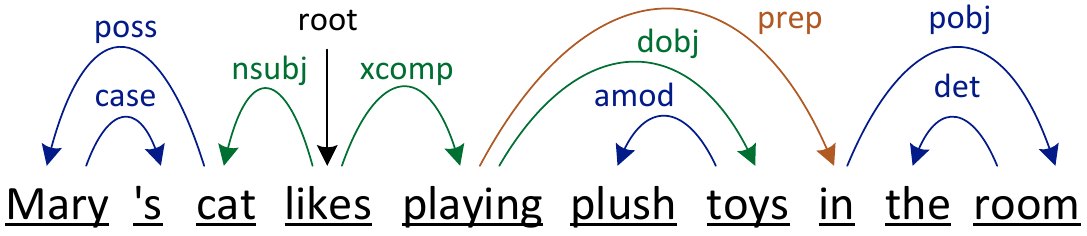}
   \caption{Dependency tree (results from spaCy).}
   \label{fig:dep_tree}
\end{subfigure}
\vspace{-1.5em}
\caption{An example of constituency tree, dependency tree, and $n$-ary OpenIE tuple to be extracted for sentence \say{\textit{Mary's cat likes playing plush toys in the room.}}}
\label{fig:syc_sentence}
\end{figure}

In general, traditional OpenIE systems are either statistical or rule-based. They extract relational tuples mainly based on certain sentence patterns heuristically defined on syntactic structures. 
The limitation of patterns causes traditional OpenIE systems to be ineffective in handling complex sentences.
Recently, neural OpenIE systems have been developed and showed promising results. Neural OpenIE systems no longer depend on pre-defined patterns.
Instead, they learn to extract relational tuples directly from unstructured text in an end-to-end manner.
However, utilizing syntactic information is under-explored among the neural OpenIE systems, although syntactic information is widely explored in other Information Extraction tasks such as Semantic Role Labeling (SRL)~\cite{fei-etal-2021-better} and Relation Extraction (RE)~\cite{zhang-etal-2018-graph}.

As syntactic information was proved to be useful for the traditional OpenIE systems, we argue that it is important for neural OpenIE systems as well. 
Figure 1a shows an $n$-ary tuple expected from an example sentence. 
Figure~\ref{fig:const_tree} displays the sentence's constituency tree, where the phrases are labeled with constituency tags.
The dependency tree (shown in Figure~\ref{fig:dep_tree}) represents syntax through directed and typed edges between words, instead of between phrases.
We observe that the boundary of the OpenIE tuple (shown in Figure~\ref{fig:tuple_extraction}) highly coincides with edges from constituency and dependency trees.
As such, we believe both constituency and dependency trees provide useful and complementary syntactic information for OpenIE systems.
The next question is: how to effectively incorporate syntactic information to neural OpenIE models?

To fully explore the syntactic information from both parse trees, we convert them into `node-sharing' graphs, \ie \textit{constituency graph} (denoted as const-graph) and \textit{dependency graph} (denoted as dep-graph), respectively. 
A dependency relation specifies a relationship between two words, and we can represent words as nodes and the relationship as corresponding edge. The key challenge is to map constituency tree to a graph in which all of its nodes are words, not phrases. Meanwhile, the graph needs to largely capture the constituency syntactic information. 
In this work, we present a novel method for the conversion of constituency trees.
With the converted const-graph modelled at the word level, we can now easily integrate it with dep-graph. Moreover, both graphs can be directly integrated with Pre-trained Language Model (PLM) which provides word-level representation.

In order to leverage heterogeneous syntactic information from both const-graph and dep-graph, we propose a novel neural OpenIE model: \mname  (\textbf{S}yntactic \textbf{M}ult\textbf{i}-view \textbf{Le}arning for \textbf{O}pen \textbf{I}nformation \textbf{E}xtraction). 
It first encodes a sentence using BERT~\cite{devlin2018bert}, and subsequently uses two syntactic encoders, namely Const-encoder and Dep-encoder. The model represents constituency and dependency relations of the sentence with the corresponding BERT representations and applies two Graph Convolutional Networks (GCN) (denoted as Const-GCN and Dep-GCN) to learn graph representations for const-graph and dep-graph separately. The representations from BERT, Const-GCN, and Dep-GCN are aggregated and finally used for tuple generation.
To better fuse the heterogeneous syntactic graph representations, \mname introduces a subtask, multi-view learning, to learn multiple types of relationships among const-graph and dep-graph. The multi-view learning loss is used to finetune the graph representations along with OpenIE loss. In summary, our contributions are threefold:
\setlist{nolistsep}
\begin{itemize}[noitemsep,leftmargin=5.5mm]
  \item We propose a novel strategy to map phrase-level relations in constituency tree into word-level relations, and to enhance each word's representation with constituency path information.
  \item We propose \textbf{\mname}, the first neural OpenIE system that incorporates heterogeneous syntactic information through GCN encoders and multi-view learning.
  \item Our experimental results show that the proposed neural OpenIE model achieves better performance than state-of-the-art methods.
\end{itemize}

%======================================================================
\section{Related Work}
%======================================================================

\paragraph{Syntax Usage in OpenIE.}
Open Information Extraction (OpenIE) was first proposed by \citet{yates2007textrunner}, and TextRunner is the first OpenIE system that generates relational tuples in open domain. 
Before deep learning era, many statistical and rule-based systems have been proposed, including Reverb~\cite{fader2011identifying}, R2A2~\cite{fader2011identifying}, OLLIE~\cite{schmitz2012open}, Clausie~\cite{del2013clausie}, Stanford OpenIE~\cite{angeli2015leveraging}, OpenIE4~\cite{mausam2016open}, NESTIE~\cite{bhutani2016nested}, and MINIE~\cite{gashteovski2017minie}. 
Most of these models extract relational tuples based on syntactic structures such as part-of-speech (POS) tags and dependency trees.
In this sense, syntactic information has been essential to OpenIE.

Recently, neural OpenIE systems~\cite{cui2018neural, stanovsky2018supervised, roy2019supervising, kolluru2020imojie, dong2021docoie, vasilkovsky2022detie, kotnis-etal-2022-milie} have been developed and showed promising results. 
Neural OpenIE systems are able to extract relational tuples end-to-end based on the semantic encoding of input sentence.
The analysis of syntactic structure of sentence, which was required by traditional models, seems no longer necessary.
As a result, the usage of syntactic information is under-explored in neural OpenIE models. 
Nevertheless, there exist some neural OpenIE systems that utilize some forms of syntactic information.
For example, RnnOIE~\cite{stanovsky2018supervised} projects POS tag of each word into POS embedding and concatenates it with word embedding as input to sentence encoder. 
SenseOIE~\cite{roy2019supervising} further concatenates word embedding with dependency embedding.
CIGL-OIE~\cite{kolluru2020openie6} finds all head verbs in the sentence and pre-defines a few POS patterns to explicitly constrain the model training. 
MGD-GNN~\cite{MGD-GNN_2021} connects words, if they are in dependency relations, in an undirected graph and applies graph attention network (GAT) to the graph~\cite{velivckovic2017graph} as its graph encoder. Although MGD-GNN uses some graphic information of dependency, %however, 
it loses other information like the directness and types of dependency relations.

In short, we observe that existing neural OpenIE systems fail to explore some syntactic features and that the integration of syntax is in a shallow manner.
Compared to the them, our \mname is able to leverage full features of heterogeneous syntactic information from both constituency and dependency trees.

\paragraph{Integration of Constituency and Dependency Syntax.}
Although constituency and dependency trees possess common sentential syntactic information, they capture syntactic information from different perspectives. Recent NLP tasks have benefited from integrating these two syntactic representations.
\citet{zhou-zhao-2019-head} and \citet{strzyz-etal-2019-sequence} integrate dependency and constituency syntactic information as a representation of parse tree or sequence, but not of a graph.
To the best of our knowledge, HeSyFu~\cite{fei-etal-2021-better} is the only work that converts dependency and constituency trees into graphs and performs graph learning strategy on both.
In this sense, although HeSyFu is designed for SRL task, it is the most relevant model to ours.

\mname differs from HeSyFu mainly in three perspectives:
\textbf{(1)} HeSyFu models constituency tree at phrase level, which is inconsistent with word-level representations from BERT. Meanwhile, HeSyFu models dependency tree at word level, so the constituency representations cannot be directly fused with the dependency representations. \textbf{(2)} As a result, HeSyFu integrates the BERT representations and the two parse trees' representations with complicated bridging processes, which may hinder synergistic integration of the heterogeneous representations.
\textbf{(3)} To better fuse heterogeneous syntactic information, \mname adopts multi-view learning to capture multiple relationships between const-graph and dep-graph representations.

\paragraph{Multi-view Learning}

Multi-view learning aims to learn representations or features from the multi-view data. Generally, data from different views usually contain complementary information. Therefore, multi-view learning is able to exploit such complementary information to learn more comprehensive representations than those of single-view learning methods~\cite{8471216}.
In the modern era, multi-view data have increased voluminously, leading to more attention of multi-view learning mechanism. The studies of multi-view learning~\cite{YAN2021106} mainly fall into: multi-view fusion~\cite{ZHAO201743,sun2013survey}, multi-modal learning~\cite{8103116,8269806}, multi-view clustering~\cite{9395530}, and multi-view representation learning~\cite{8471216,8715409, mane_multi_view_2021}.
In our work, we perform multi-view learning and fusion on two views of syntactic graphs ( \ie const-graph and dep-graph).
To the best of our knowledge, we are the first to use multi-view mechanism to exploit complementary syntactic information in NLP applications.

%======================================================================
\section{Graph Modelling}
%======================================================================
In this section, we elaborate on our graph modelling strategy to convert constituency and dependency trees into graphs $G=(U,E)$, where $U$ indicates the set of nodes and $E$ the set of edges.
They are $G^{con}=(U^{con},E^{con})$ for const-graph, and $G^{dep}=(U^{dep},E^{dep})$ for dep-graph. The two graphs' nodes correspond to the same set of input sentence's words. 
However, the labels of the nodes in the two graphs are different (constituency path for const-graph, and dependency relation type for dep-graph), preserving the syntactic information of constituency and dependency trees. Also, their edge sets are different, where the edges represent word-to-word syntactic relations.

%===========================================================
\subsection{Dependency Graph Modelling}
%===========================================================

\begin{figure}
    \centering
    \includegraphics[width=0.8\columnwidth]{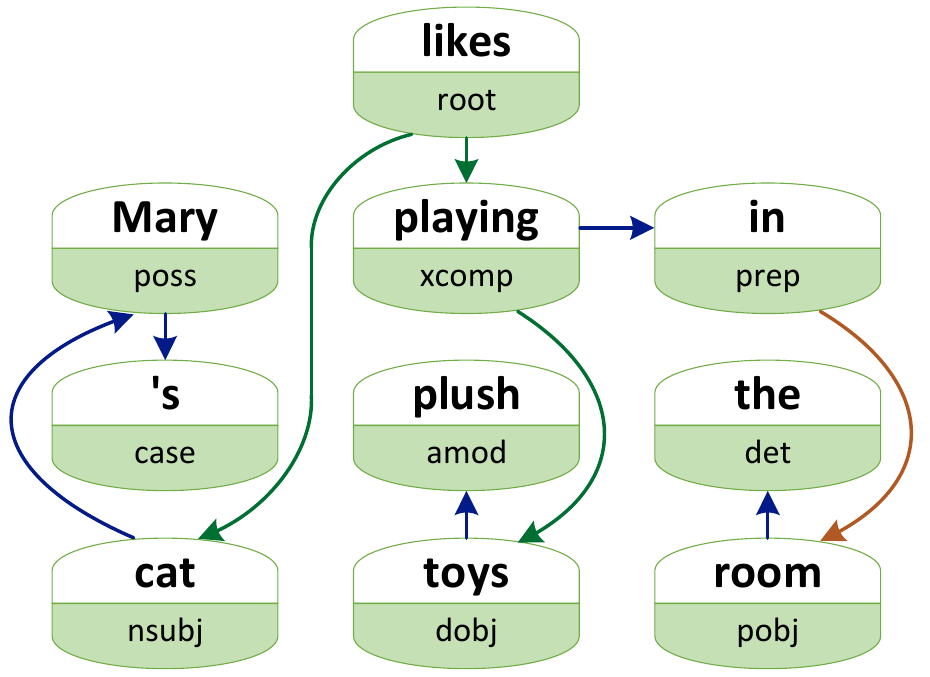}
    \vspace{-0.5em}
    \caption{Dependency graph constructed from word-level dependency relations in Figure~\ref{fig:syc_sentence}.}
    \label{fig:dep_graph}
    \vspace{-1.5em}
\end{figure}

Dependency tree provides syntactic dependency at word level. Thus, the dep-graph of a sentence is identical to the dependency tree of the sentence, except node labels.
For each word, as shown in Figure~\ref{fig:dep_tree}, there is an inbound relation from its modifying head word. 
We follow \citet{fei-etal-2021-better} to label each word node with its inbound dependency relation type, as exemplified in Figure~\ref{fig:dep_graph}.

%===========================================================
\subsection{Constituency Graph Modelling}\label{sec:const_graph}
%===========================================================
We flatten the phrase-level relations of a constituent structure into a const-graph of word-level relations, which can be directly integrated with word-level granularity from Pre-trained Language Model (PLM) such as BERT. 
The flattening process is designed to preserve both the phrasal boundary information and the constituency relations which are required for OpenIE task. But note that this flattening process can be used for other related modelling tasks (\eg SRL, NER, RE).

\begin{figure}
    \centering
    \includegraphics[width=1\linewidth]{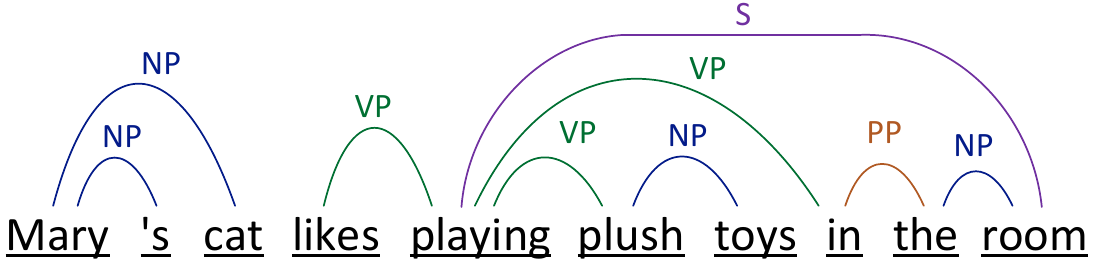}
    \caption{Word-level constituency relations converted from phrase-level relations of constituency tree.}
    \label{fig:const_arc}
\end{figure}

\begin{table}[t]
\small
    \begin{tabular}{c|l}
    \toprule
    Word & Const. Path \\
    \midrule
    Mary & S-NP-NP  \\
    `s & S-NP-NP  \\
    cat & S-NP  \\
    likes & S-VP  \\
    playing & S-VP-S-VP  \\
    \bottomrule
    \end{tabular}
    \quad
    \begin{tabular}{c|l}
    \toprule
    Word & Const. Path \\
    \midrule
    plush & S-VP-S-VP-NP  \\
    toys & S-VP-S-VP-NP  \\
    in & S-VP-S-VP-PP  \\
    the & S-VP-S-VP-PP-NP  \\
    room & S-VP-S-VP-PP-NP \\
    \bottomrule
    \end{tabular}
\caption{Example constituency paths.}
\label{tab:const_path}
\end{table}

\paragraph{Word Node Labelling with Constituency Path.}

In const-graph, each word is a node, and we label each word node with the path from the root to the word in the constituency structure of the input sentence. Table \ref{tab:const_path} lists the constituency paths of words in the example sentence in Figure \ref{fig:syc_sentence}. This labelling of words with constituency paths preserves the rich phrasal information of constituency tree.

\paragraph{Word-level Constituency Relations.}

In const-graph, edges are constituency relations that connect word nodes.
We perform relation flattening of the constituency tree in Figure \ref{fig:const_tree} in the following steps:  
\textbf{(1)} We add an edge between the first and last word in each noun phrase (NP) (\eg `Mary'-`'s', `Mary'-`cat', `plush'-`toys', `the'-`room'). The edge is labelled as `NP'. This edge identifies the boundary of NP.
\textbf{(2)} If a word and a phrase are siblings (belonging to the same parent node in constituency tree), we connect the word (\eg Verb in VP, Preposition in PP) to the first word of its sibling phrase (\eg `likes'-`playing', `playing'-`plush', `playing'-`in', `in'-`the').
This edge's type is marked according to the constituency type of the parent node such as VP and PP.
\textbf{(3)} To mark the boundary of intra-sentential clause, we connect the first and last word of each clause (S) and label the edge as `S' (\eg `playing'-`room'). 
\textbf{(4)} We remove an edge when the distance between two words is longer than 8 in the input sentence, since the elements in OpenIE tuple are usually short spans.

\begin{figure}
   \centering
   \includegraphics[width=0.8\linewidth]{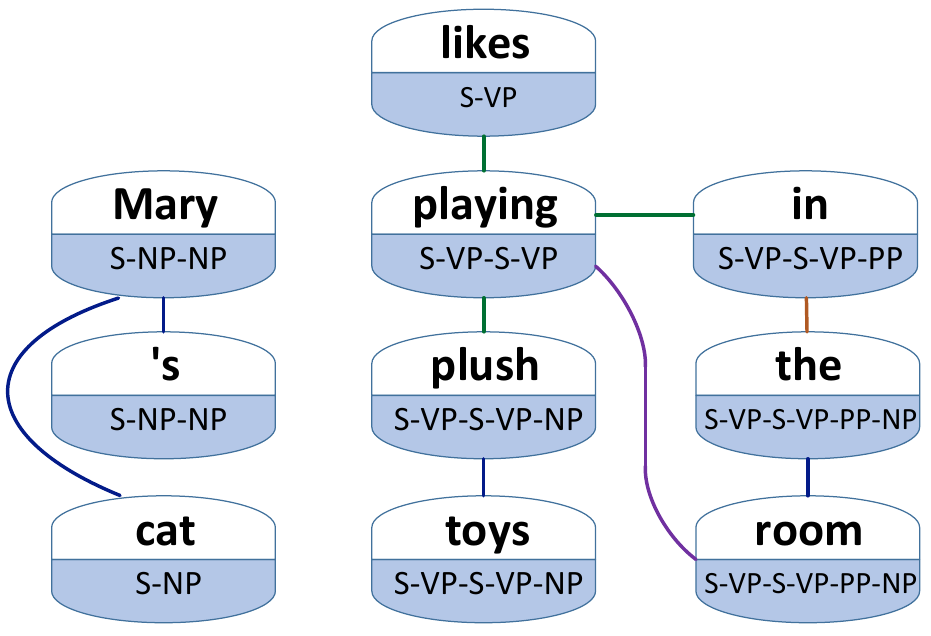}
   \caption{Constituency graph constructed from word-level constituency relations in Figure~\ref{fig:const_arc} and hierarchical constituency paths in Table~\ref{tab:const_path}.}
   \label{fig:const_graph}
\end{figure}

Figure~\ref{fig:const_arc} depicts the word-level relations flattened from the constituency tree in Figure~\ref{fig:const_tree}, and Figure~\ref{fig:const_graph} depicts the final const-graph.

%======================================================================
\section{\mname Model}\label{sec:model}
%======================================================================
\setlength{\abovedisplayskip}{3pt}
\setlength{\belowdisplayskip}{3pt}

\begin{figure*}
   \centering
   \includegraphics[width=0.95\linewidth]{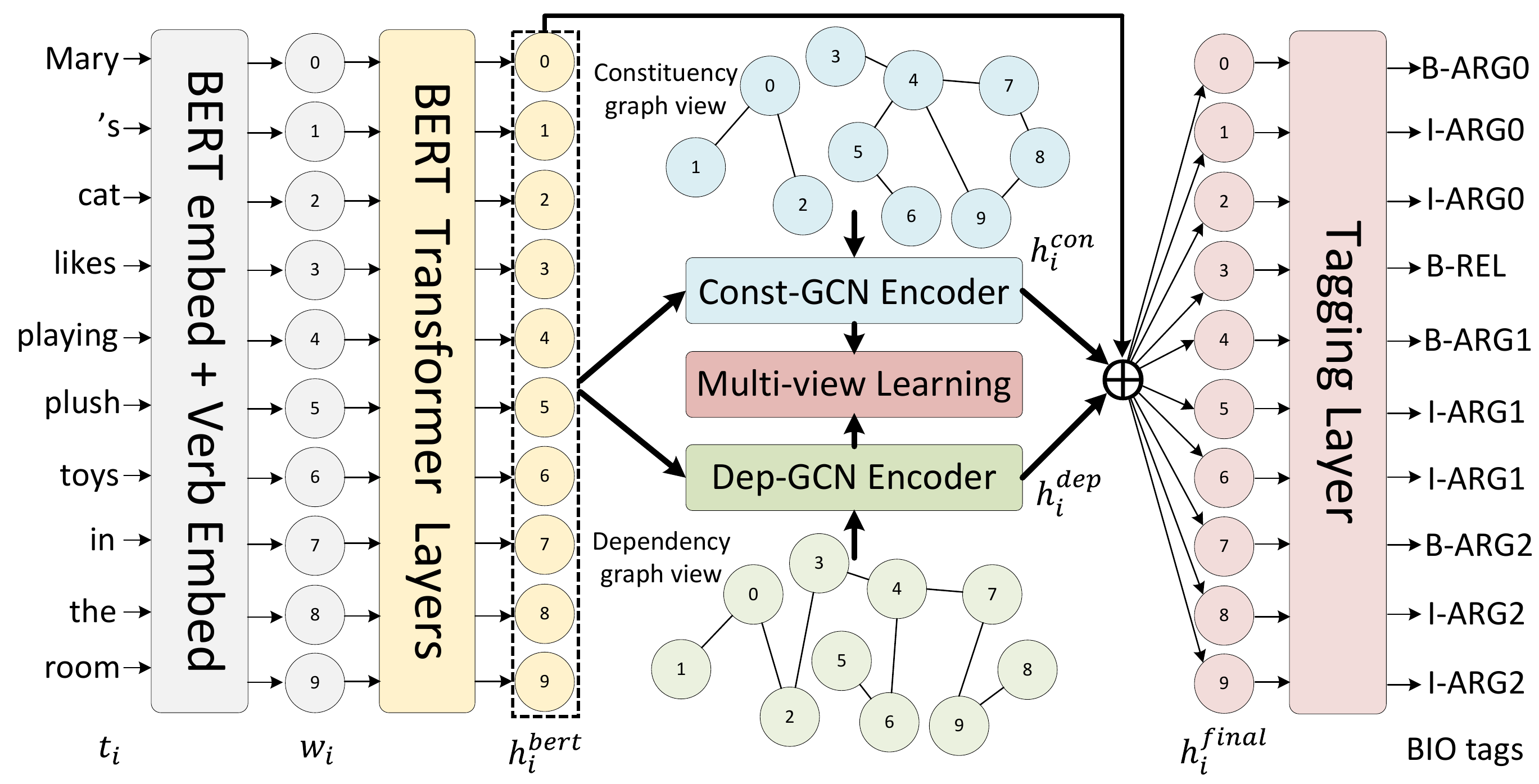}
%   \vspace{-1.0em}
   \caption{Overview of \mname model architecture.}
   \label{fig:model_diagram}
\end{figure*}

The overall architecture of \mname is illustrated %displayed 
in Figure~\ref{fig:model_diagram}. \mname is based on BERT encoder to get contextualized representations of an input sentence. The BERT representations are then integrated with constituency and dependency information by Const-GCN and Dep-GCN, respectively. Finally, \mname aggregates the BERT, const-graph, and dep-graph representations in order to predict output tuples. Beyond the OpenIE tagging loss, \mname  further performs multi-view learning on the const-graph and dep-graph representations, and generates additional multi-view losses to enhance OpenIE tagging accuracy.

%===========================================================
\subsection{Task Formulation}
%===========================================================

We formulate OpenIE as a sequence tagging task, using BIO (Beginning, Inside, Outside) tagging scheme like recent neural OpenIE models~\cite{stanovsky2018supervised}.
Given an input sentence $s=[t_0, \dots, t_n]$, a variable number of tuples will be extracted. Each tuple can be represented as $[x_0, \dots, x_m]$, where $x_j$ is a contiguous subspan of $s$.
One of $x_j$ is tuple relation (B/I-REL) and the others are tuple arguments (B/I-ARG$_l$).

We assume that each tuple has a verb as relation, since OpenIE relation is typically associated with a verb\footnote{Relation can be referred as predicate, and verb can be referred as predicate head word in other OpenIE works.}. Meanwhile, since some verbs in a sentence do not lead to any relational tuple, we assume one verb can be a relation for at most one tuple.

%===========================================================
\subsection{BERT Encoder with Relation Indicator}\label{subsec:bert_encoder}
%===========================================================

We employ BERT~\cite{devlin2018bert} as our encoder to analyze semantic interactions among words.
We first project all words $[t_0, \dots, t_n]$ into embedding space by summing their word embedding\footnote{If the word contains multiple sub-words after BERT tokenization, we use the representation of its first sub-word.} and verb embedding, \ie $w_i=W_{word}(t_i)+W_{verb}(t_i)$.
Here, $W_{word}$ is trainable and initialized by BERT word embedding.

$W_{verb}$ is a trainable verb embedding matrix. 
Verb embedding is to distinguish whether an input word is a relation indicator or not.
Given an input sentence, we extract all verbs from the sentence using an off-the-shelf POS tagger.
We consider each verb in a sentence to be a potential relation indicator and use the verb embedding to highlight this relation indicator. 
Specifically, $W_{verb}$ initializes each verb to 1 at a time, and all the other words in the sentence to 0.
If a verb in the sentence does not lead to a tuple, we set all of the sequence output tags to be \say{O}.
Consequently, the model is able to learn which verbs lead to relation.\footnote{Implementation details of verb-tuple alignment are described in Appendix \ref{sec:expSetup}.}

Then, we use $w_s=[w_0, \dots, w_n]$ as the input to the BERT encoder and utilize BERT's last hidden states as contextualized representations:
% \useshortskip
\begin{equation}\label{eq:h_i}
    h_i^{bert} = \mathrm{BERT}(w_i)  \in \mathbb{R} ^ {d_h}
\end{equation}

%===========================================================
\subsection{Syntactic GCN Encoders}
%===========================================================

In this section, we present \textit{syntactic encoders}, which represent the elements of the two graphs with the BERT representations and encode them using GCNs.
Recall that the dep-graph and const-graph are represented as $G^z=(U^z,E^z)$, where $z \in \{dep, con\}$. 
$e^z_{ij}$ in $E^z$ equals to 1 if there is an edge between node $n^z_i$ and node $n^z_j$;
Otherwise, 0. Each node $n^z_i \in U^z$ has a label (or type), designated as $type\left<n_i^z\right>$.

The node types of $U^{dep}$ are dependency relations. 
The  syntactic encoder of $G^{dep}$ (called Dep-GCN) takes node embedding as follows:
\begin{equation}
    l^{dep}_{i}=W_{dep}^1\big(type\big<n_i^{dep}\big>\big)
\end{equation}
where $W_{dep}^1 \in \mathbb{R} ^ {d_l \times N_{dep}}$ is a trainable matrix, $d_l$ is the size of node embeddings, and $N_{dep}$ is the total number of unique dependency relations. 

A node $n^{con}_i \in U^{con}$ has a label of constituency path $type\left<n_i^{con}\right>$ = $[n_i^{con_0}, \cdots, n_i^{con_m}]$, which contains a list of constituent tags. Given a constituency path, we first project all its constituent tags to respective constituent tag vectors, and then average all the constituent tag vectors in this constituency path as inputs to the syntactic encoder of $G^{con}$ (called Const-GCN) as follows:
\begin{align}
    l^{con}_{i} &= \mathrm{avg} \big( W_{con}^1(type\left<n_i^{con}\right>) \big) \\
    &= \mathrm{avg} \big(W_{con}^1(n_i^{con_0}), \cdots, W_{con}^1(n_i^{con_m}) \big) \notag
\end{align}
where $W_{con}^1 \in \mathbb{R} ^ {d_l \times N_{con}}$ is a trainable matrix, and $N_{con}$ is the total number of unique constituency tags. avg() indicates the averaging operation on a sequence of vectors.

Each syntactic encoder (Dep-GCN, Const-GCN) employs a separate GCN to encode the corresponding graph ($G^{dep}$, $G^{con}$). The computation of the GCN representation is formulated as:
\begin{equation}\label{eq:h_dep}
    h_i^{z} = \mathrm{ReLU}\Big(\sum_{j=1}^{n} \alpha^{z}_{ij}(h_j^{bert} + W_z^2 \cdot l^{z}_{j} + b_z) \Big)
\end{equation}
where $n$ refers to the total number of word nodes in the graph, $W_z^2  \in \mathbb{R} ^ {d_h \times d_l}$ is a trainable weight matrix for syntactic type embeddings, and $b_z  \in \mathbb{R} ^ {d_h}$ is the bias vector. The neighbour connecting strength distribution $\alpha^{z}_{ij}$ is calculated as below:
\begin{equation}
    \alpha^{z}_{ij} = \frac{e^{z}_{ij} \cdot \mathrm{exp}\big( (m_i^{z})^T \cdot m_j^{z} \big)}{\sum_{k=1}^{n}e^{z}_{ik} \cdot \mathrm{exp} \big( (m_i^{z})^T \cdot m_k^{z} \big)}
\end{equation}
% , where $m_i^{z} = h_i^{bert} \oplus l^{z}_{i} \in \mathbb{R} ^ {d_h + d_l}$.
where $m_i^{z} = h_i^{bert} \oplus l^{z}_{i}$, and $\oplus$ is concatenation operator.
In this way, node type and edge information are modelled in a unified way.

Finally, we aggregate the sequence representations from BERT encoder in Eq.(\ref{eq:h_i}) and the graph representations from Const-Encoder and Dep-Encoder in Eq.(\ref{eq:h_dep}) as follows:
\begin{equation} 
    h_i^{final} =  h_i^{bert} \oplus h_i^{con} \oplus h_i^{dep}
\end{equation}
where $h_i^{final}$ is used by the tagging layer for tuple prediction.

%===========================================================
\subsection{Multi-view Learning}
%===========================================================

Recall that the const-graph and the dep-graph share the same set of nodes $U$ and have two different sets of node representations $h^{con}$ and $h^{dep}$, and two different edge sets $E^{con}$ and $E^{dep}$, respectively. We treat const-graph and dep-graph as two syntactic views $z \in \{dep,con\}$ of the input sentence.
We adopt multi-view learning~\cite{mane_multi_view_2021} in order to explore three types of relationships among the representations of const-graph and dep-graph views. The multi-view learning loss is used to finetune these representations, which can provide rich syntactic information for tuple generation.

We consider three categories of relationships between these two views. In the first category, the multi-view learning captures the inter-node and intra-view relationship in each view. In the second category, it aligns instances of the same node across various views, \ie intra-node and inter-view relationship. In the third category, it ensures the nodes that are connected in one view should be similar with each other in another view, \ie inter-node inter-view relationship.

\paragraph{Inter-node Intra-view Relationship.}
We design a loss to ensure the representations of connected nodes $i$ and $j$ in the same view $z$, \ie $h_i^z$ and $h_j^z$, to be similar.\footnote{In the original paper~\cite{mane_multi_view_2021}, the corresponding objective of their multi-view learning is to make node representations within each view close to each other. In our problem setting, we redefine this objective on `connected nodes' only.} This is to ensure coherence within the same view.
\begin{align}
    &L_{R_1} = - \sum_{z \in \{dep, con\}} \sum_{i \in U} \sum_{j \in U} e_{ij}^z \cdot \mathrm{log} P(h_j^z ,h_i^z) \label{eq:loss_div} \\
    &P(h_j^z ,h_i^z) =  \frac{\mathrm{exp}(h_j^z \cdot h_i^z)}{\sum_{k \in U} \mathrm{exp} (h_k^z \cdot h_i^z)} \label{eq:softmax}
\end{align}

\paragraph{Intra-node Inter-view Relationship.}
While const-graph and dep-graph exhibit diversity, they ultimately converge on a common set of words.
The same word, although bearing different syntactic functions, well connects the two views.
Therefore, we design a loss for the intra-node and inter-view relations. Specifically, we make sure a node $i$'s const-graph representation $h_i^{con}$ to be similar to its dep-graph representation $h_i^{dep}$ by minimizing the following loss:
\begin{equation}\label{eq:loss_c1}
    L_{R_2} = - \sum_{z\in \{dep,con\}} \sum_{i \in U} \sum_{z' \not= z}  \mathrm{log}  P(h_i^{z'} ,h_i^z)
\end{equation}
where $z'$ indicates the other view than $z$, and $ P(h_i^{z'} ,h_i^z)$ is computed in a similar way as in Eq.~\ref{eq:softmax}.

\paragraph{Inter-node Inter-view Relationship.}
We observe that const-graph and dep-graph share many common edges. In another word, two nodes linked in const-graph are often linked with each other in dep-graph as well.
Consequently, we explore inter-view and inter-node relations in order to leverage the frequent edge sharing between the two graphs.
Specifically, if node $i$ and node $j$ are connected in const-graph, we design a loss to move node $i$`s const-graph representation $h_i^{con}$ towards node $j$`s dep-graph representation $h_j^{dep}$, as follows:
\begin{align}\label{eq:loss_c2}
    L_{R_3} &= - \sum_{z \in \{dep, con\}} \sum_{i \in U} \sum_{z' \not= z} \sum_{j \in U} e_{ij}^z \cdot \mathrm{log} P(h_i^{z'} , h_j^z)
\end{align}

\paragraph{Loss Function.}
We combine the losses of the three categories of relations in Equations (\ref{eq:loss_div}), (\ref{eq:loss_c1}), and (\ref{eq:loss_c2}) with the OpenIE sequence tagging loss $L_{CE}$. $L_{CE}$ is the cross-entropy loss between the gold and the predicted word labels in sequence tagging (\ie the BIO tags shown in Figure~\ref{fig:model_diagram}). The overall loss for our multi-view learning OpenIE is:
\begin{equation}
 L = L_{CE} + \alpha \cdot L_{R_1} + \beta \cdot  L_{R_2} + \gamma \cdot L_{R_3}
\end{equation}
where  $\alpha$, $\beta$, and $\gamma$ are hyper-parameters, indicating the importance of each individual loss. 

%======================================================================
\section{Experiments}
%======================================================================

\begin{table}[t]
\small
\centering
\begin{tabular}{ l|crr}
 \toprule
 Dataset & Source & \#Sent & \#Tuple\\
 \midrule
 LSOIE-wiki-train & QA-SRL 2.0 & 19,591 & 45,890 \\
 LSOIE-wiki-test & QA-SRL 2.0 & 4,660 & 10,604 \\
 \midrule
 LSOIE-sci-train & QA-SRL 2.0 & 38,826 & 80,271 \\
 LSOIE-sci-test & QA-SRL 2.0 & 9,093 & 17,031 \\
  \midrule
 CaRB-train & OpenIE 4 & 92,774 & 190,661 \\
 CaRB-test & Crowdsourcing & 1,282 & 5,263 \\
 \bottomrule
\end{tabular}
\vspace{-0.5em}
\caption{Statistics of OpenIE datasets used in training and evaluating \mname.}
\vspace{0.5em}
\label{tab:data_set}
\end{table}

\begin{table*}
\centering
\begin{tabular}{l|ll|ll|ll}
 \toprule
 \multirow{2}{*}{Models} &
 \multicolumn{2}{c|}{LSOIE-wiki} &
 \multicolumn{2}{c|}{LSOIE-sci} &
 \multicolumn{2}{c}{CaRB} \\
  & F1 & AUC & F1 & AUC & F1 & AUC\\
 \midrule
 \textbf{Without Dependency \& Constituency Graph} & & & & & &\\
  GloVe + bi-LSTM~\cite{stanovsky2018supervised} & 43.90 & 38.04       & 50.51 & 45.95        & 48.52 & 27.10\\
  GloVe + bi-LSTM + CRF & 44.48 & 38.72         & 50.85 & 46.23        & 48.72 & 27.51\\
  BERT~\cite{solawetz-larson-2021-lsoie}  & 47.54 & 44.71         & 57.02 & 53.23        & 51.45 & 30.62\\
  CopyAttention~\cite{cui2018neural} & 39.52 & 35.99 & 48.82 & 46.84 & 51.6$^\dag$ & 32.8$^\dag$ \\
  IMoJIE~\cite{kolluru2020imojie}  & 49.24 & 47.55 & 58.75 & 55.81 & 53.5$^\dag$ & 33.3$^\dag$ \\
  CIGL-OIE + IGL-CA~\cite{kolluru2020openie6} & 44.75 & 41.98 & 56.62 & 52.37 & \textbf{54.0}$^\dag$ & \textbf{35.7}$^\dag$ \\
 \midrule
 \textbf{With Dependency Graph} & & & & & &\\
 BERT + Dep-GCN & 48.71 & 47.87        & 58.14 & 55.32        & 52.49 & 32.85\\

 \midrule
 \textbf{With Constituency Graph} & & & & & &\\
 BERT + Const-GCN & 49.74 & 48.55         & 58.67 & 55.76        & 52.83 & 33.10\\
 
 \midrule
 \textbf{With Dependency \& Constituency Graph} & & & & & &\\
 BERT + Dep-GCN $\oplus$ Const-GCN & 49.89 & 49.15        & 59.23 &  \underline{56.47}        & 52.70 & 33.28\\
 BERT + Dep-GCN $\rightarrow$ Const-GCN &  \underline{50.21} &  \underline{49.28}       &  \underline{59.53} & 55.92        & 52.85 & 33.37\\
 BERT + Const-GCN $\rightarrow$ Dep-GCN & 49.71 & 48.80         & 58.81 & 56.04       & 53.28 & 33.51\\
 \mname & \textbf{51.73} & \textbf{50.88} & \textbf{60.51} & \textbf{57.22} & \underline{53.76} & \underline{34.92} \\

 \bottomrule
\end{tabular}
\caption{Results on OpenIE datasets. Scores with $^\dag$ are from \citet{kolluru2020openie6}. The best scores are in boldface, and the second best scores underlined.}

\label{tab:baseline_systems}
\end{table*}

We mainly conduct our experiments on LSOIE~\cite{solawetz-larson-2021-lsoie}, a large-scale OpenIE data converted from QA-SRL 2.0 in two domains, i.e., Wikipedia and Science. It is 20 times larger than the next largest human-annotated OpenIE data, and thus is reliable for fair evaluation.\footnote{https://github.com/Jacobsolawetz/large-scale-oie}
LSOIE provides $n$-ary OpenIE annotations and gold tuples are in the $\langle ARG_0, Relation, ARG_1, \dots, ARG_n \rangle$ format.
The dataset has two subsets, and we use both, namely LSOIE-wiki and LSOIE-sci, for comprehensive evaluation. LSOIE-wiki has 24,251 sentences and LSOIE-sci has 47,919 sentences.

CaRB~\cite{bhardwaj2019carb} dataset is the largest crowdsourced OpenIE dataset.\footnote{https://github.com/dair-iitd/CaRB} 
However, CaRB only provides 1,282 annotated sentences, which are insufficient for training neural OpenIE models.
As a result, we use the CaRB dataset purely for testing. We follow \citet{kolluru2020openie6} to convert bootstrapped OpenIE4 tuples as labels for distant supervised model training.
CaRB provides binary OpenIE annotations and gold tuples are in the form of $\langle Subject, Relation, Object \rangle$.
Finally, we summarize the statistics of the training and testing datasets of LSOIE-wiki, LSOIE-sci, and CaRB in Table~\ref{tab:data_set}.

\subsection{Baselines for Comparison}
\paragraph{Baselines without Syntax.} RnnOIE~\cite{stanovsky2018supervised} is the first sequence tagging model based on Bi-LSTM networks. In our work, we implement\footnote{Details of implementation are described in Appendix~\ref{sec:reimplementation}.} its model with GloVe word representation~\cite{pennington-etal-2014-glove}, and name it as `GloVe+bi-LSTM'. We further add a CRF layer to be another baseline model `GloVe+bi-LSTM+CRF'. Meanwhile, we utilize BERT word representation along with its transformer layers, and name this baseline model as `BERT'. `CopyAttention'~\cite{cui2018neural} is the first neural OpenIE model which casts tuple generation as a sequence generation task. `IMOJIE'~\cite{kolluru2020imojie} extends CopyAttention and is able to produce a variable number of extractions per sentence. It iteratively generates the next tuple, conditioned on all previously generated tuples. `CIGL-OIE + IGL-CA'~\cite{kolluru2020openie6} models OpenIE as a 2-D grid sequence tagging task and iteratively tags the input sentence until the number of extractions reaches a pre-defined maximum.

\paragraph{Baselines with Syntax.}
We build baselines that utilize syntactic information, based on BERT. We first study the performance of using either dependency or constituency tree as additional syntactic feature, \ie `BERT+Dep-GCN' and `BERT+Const-GCN'. Then, we present three models of fusing heterogeneous syntactic information from dependency and constituency trees. `Dep-GCN $\oplus$ Const-GCN' refers to the proposed parallel aggregation of the two graph representations using two syntactic GCNs. For comparison, we build a model of sequential aggregation `Dep-GCN $\rightarrow$ Const-GCN', which passes the dependency graph representation from Dep-GCN as input to Const-GCN, and another model `Const-GCN $\rightarrow$ Dep-GCN', which passes the constituency graph representation from Const-GCN as input to Dep-GCN.\footnote{Details of experimental setups and re-implementation details are described in Appendix \ref{sec:expSetup} and \ref{sec:reimplementation}.}

\subsection{Evaluation}

\paragraph{Evaluation Metric.}

For LSOIE-wiki and LSOIE-sci dataset, \citet{solawetz-larson-2021-lsoie} consider two tuples to match if their relations (or verbs) are identical, regardless of the matching of tuple arguments. We consider this scoring function to be over-lenient. Therefore, we revise their scoring function to consider both relation and arguments matching, \ie exact tuple matching, for accurate and fair comparison.
For CaRB dataset, we use the default CaRB scoring function~\cite{bhardwaj2019carb} to evaluate binary tuple with lexical level matching, \ie partial tuple matching.
Both scoring functions report F1 score based on precision and recall computed by tuples matching. Each tuple extracted is associated with a confidence value, so we can generate a precision-recall (P-R) curve and report the area under P-R curve (AUC).

\begin{figure}[t]
  \centering
  \includegraphics[width=1\linewidth]{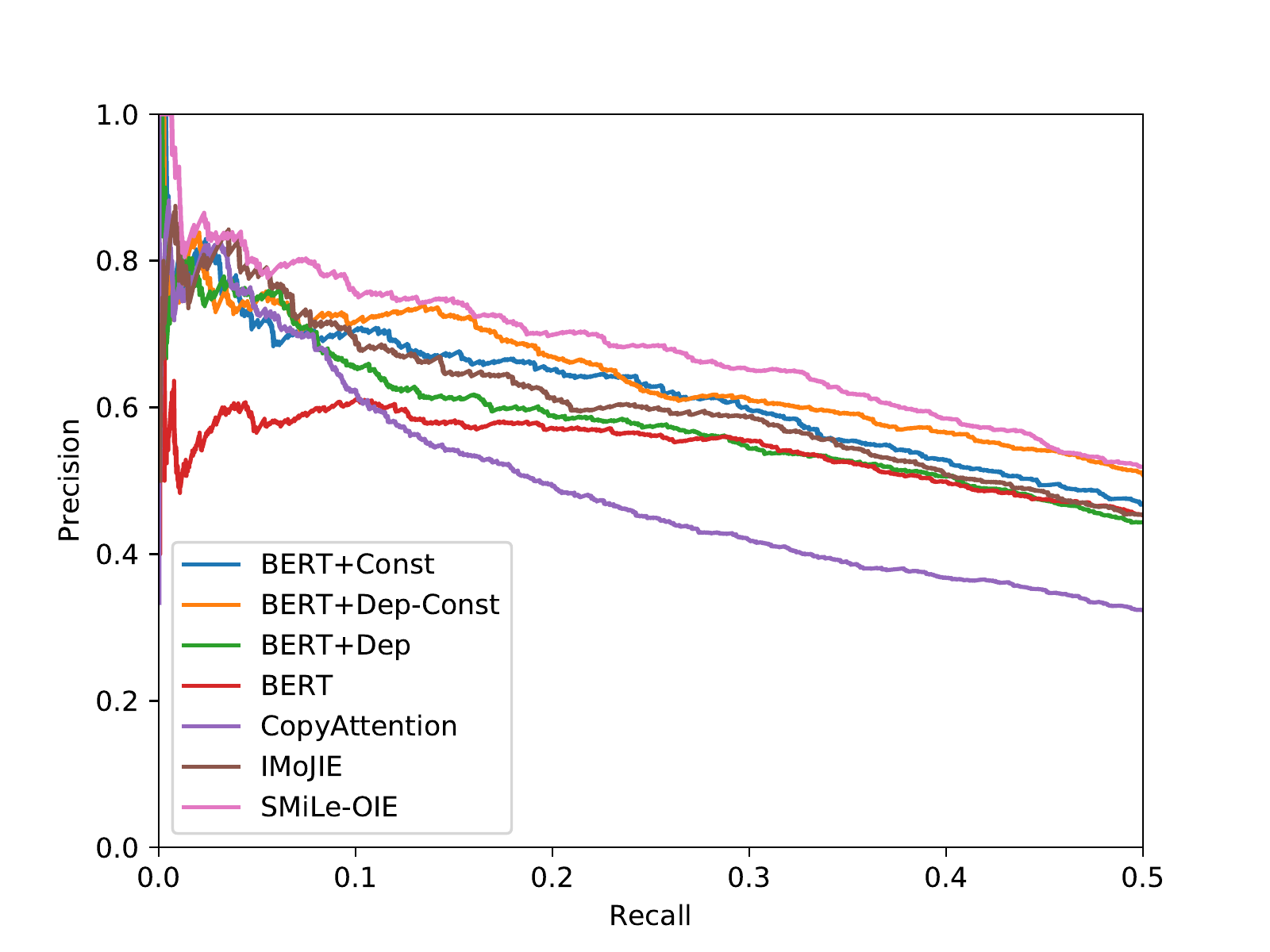}
  \vspace{-2.0em}
  \caption{Precision\jj{-}recall curves of \mname and other baselines on LSOIE-wiki test set.}
  \label{fig:auc}
\end{figure}

\subsection{Dataset}

\paragraph{Evaluation Results.}
We compare \mname with other neural OpenIE baseline systems, summarizing their evaluation results in Table~\ref{tab:baseline_systems} and depicting their P-R curves in Figure~\ref{fig:auc}. As shown in Figure~\ref{fig:auc}, \mname achieves better precision at different recalls comparing to other baseline systems.
Observe that both \say{BERT + Dep-GCN} and \say{BERT + Const-GCN} outperform \say{BERT}. 
It shows that leveraging syntactic information, either dependency or constituency tree, benefits OpenIE task significantly. 

The comparison results in Table~\ref{tab:baseline_systems} also show that the integration of the \textit{heterogeneous} syntactic information is better than leveraging a \textit{single} syntactic structure. Both parallel and sequential aggregation of the two graph representations achieve better results than \say{BERT} with either \say{Dep-GCN} or \say{Const-GCN}.

Lastly, multi-view learning can effectively guide the fusion of the heterogeneous syntactic information, leading to the significant improvement
and outperforming all the baseline systems except \say{CIGL-OIE + IGL-CA} on CaRB dataset. We find that \say{CIGL-OIE + IGL-CA} uses a complicated method of coordination boundary analysis dedicated for CaRB dataset. However, the coordination boundary analysis of \say{CIGL-OIE + IGL-CA} cannot be generalized to %well for 
other datasets, e.g., LSOIE-wiki and LSOIE-sci, and it is thus not preferable.

\begin{table}
\centering
    \begin{tabular}{l|cc|cc}
    \toprule
    \multirow{2}{*}{Models} &
    \multicolumn{2}{c|}{LSOIE-wiki} &
    \multicolumn{2}{c}{LSOIE-sci}\\
      & F1 & AUC & F1 & AUC\\
    \midrule
    \mname & \textbf{51.73} & \textbf{50.88} & \textbf{60.51} & \textbf{57.22}\\
    -- w/o $L_{R_1}$  & 50.86 & 49.23         & 59.71 & 56.43  \\
    -- w/o $L_{R_2}$ & 50.24 & 49.35         & 58.47 & 55.72      \\
    -- w/o $L_{R_3}$ & 51.05 & 50.52          & 59.84  & 56.53 \\
    \midrule
    w/o GCN  & 49.57 & 47.84         & 58.68 & 55.67     \\
    -- w/o $L_{R_1}$ & 48.15 & 46.26         & 57.22  &  53.92       \\
    -- w/o $L_{R_2}$ & 48.64 & 45.59         & 56.54  &  54.59     \\
    -- w/o $L_{R_3}$ & 49.83 & 46.62         & 57.51  & 55.24        \\
    \bottomrule
    \end{tabular}
\caption{Ablation study of \mname. The best scores are in boldface}
\label{tab:ablation}
\end{table}

\subsection{Ablation Study}
We ablate each part of our model and evaluate the ablated models against the LSOIE-wiki and LSOIE-sci datasets, and the results are reported in Table~\ref{tab:ablation}. 
The upper part of the table reports the ablation study results of removing each of the three multi-view learning losses $L_{R_1}$, $L_{R_2}$, and $L_{R_3}$. It shows that $L_{R_2}$ (intra-node inter-view relationship) has slightly more contribution than the other two losses. The lower part reports the results of removing
the GCN layers for dependency and constituency graphs. In this setting, we only concatenate the syntactic label representation to each word, without leveraging the syntactic graph structure. 
We observe that the GCN layers have larger impact on \mname than multi-view loss, although both contribute to the best performance. 
Meanwhile, we study a few variants of const-graph, and the results are reported in Appendix \ref{sec:constVariant}.

%=============================================
\subsection{Effectiveness of Const-graph}
\label{sec:constVariant}
%=============================================
To verify the effectiveness of the proposed method for converting phrase-level relations of constituency tree into word-level relations of the const-graph (see Section~\ref{sec:const_graph}), we build three variants:
\begin{itemize}
    \item \textbf{Variant 1}: we replace the constituency path, as the label of word node, with the last constituency tag in the path;
\item \textbf{Variant 2}: in step 2 of constituency relations flattening, we connect a word to the last word of its sibling phrase, instead of connecting to the first word;
\item \textbf{Variant 3}: in step 4 of constituency relations flattening, we keep edges whose distance between two words is longer than 8.
We evaluate the proposed conversion method and its three variants based on our baseline model BERT+Const-GCN. Note that neither Dep-Encoder nor multi-view learning is applied in BERT+Const-GCN.
\end{itemize}

\begin{table}[t]
\centering
    \begin{tabular}{l|cc|cc}
    \toprule
    BERT+ &
    \multicolumn{2}{c|}{LSOIE-wiki} &
    \multicolumn{2}{c}{LSOIE-sci}\\
    Const-GCN & F1 & AUC & F1 & AUC\\
    \midrule
    const-graph & \textbf{49.74} & \underline{48.55}         & \textbf{58.67} & \textbf{55.76}   \\
    - variant 1 & 48.91 & 48.04         & 57.73 & 54.89   \\
    - variant 2 & \underline{49.57} & \textbf{48.67}         & 58.17 & \underline{55.32}   \\
    - variant 3 & 49.25 & 47.78         & \underline{58.25} & 54.03   \\
    \bottomrule
    \end{tabular}
\caption{Effectiveness study of const-graph: the best scores are in boldface, and the second best underlined.}
\label{tab:const-graph-effect}
\end{table}

As shown in Table~\ref{tab:const-graph-effect}, the proposed method outperforms all the three variants. 
The const-graph with constituency path outperform its variant 1 that uses a single constituency tag. It proves that using constituency path is better than simply using the last constituency tag.
Comparing to variant 2, the const-graph which connects the word to first word of its sibling, achieves better scores.
Moreover, we find the keeping distant edges in variant 3 can deteriorate the model performance. As such, it is effective to remove distant edges from const-graph.

%======================================================================
\section{Conclusion}
%======================================================================
We design a novel strategy to map constituency tree into constituency graph only with word nodes, paving way for integrating constituency syntax with BERT and dependency syntax.
With the aid of Const-GCN and Dep-GCN, we propose a new OpenIE system \mname which combines heterogeneous syntactic information through multi-view learning. Experiment results show that leveraging syntactic information can benefit OpenIE task significantly, and multi-view learning can effectively guide the heterogeneous syntactic information fusion. In future work, we will explore other types of structured information to further improve OpenIE.

\section*{Acknowledgments}
This research is supported by the Agency for Science, Technology and Research (A*STAR) under its AME Programmatic Funding Scheme (Project \#A18A2b0046 and \#A19E2b0098).

\clearpage
%======================================================================
\section*{Limitations}
%======================================================================

We analyze the limitation of our \mname from three perspectives: syntactic parse errors, POS tagging errors and multiple extractions issue.
(1) As we integrate both constituency and dependency parsing results with OpenIE task, our system will inevitably suffer from the noises introduced by the off-the-shelf tools: spaCy and CoreNLP.
(2) Meanwhile, the number of tuple extractions is highly correlated with the number of verbs extracted by the POS tagger.
Therefore, the POS tagger's errors may also affect the quality of OpenIE. Based on our statistics of LSOIE-wiki and LSOIE-sci, the POS tagger fails to extract 8\% of verbs that are supposed to be relation indicators. 
(3) Moreover, there are 6\% of relation indicators corresponding to multiple tuple extractions, while our system extracts up to one tuple per relation indicator. 
Our system, suffering from the POS errors and the multiple extractions issue, fails to predict 14\% of the gold tuples.

\bibliography{anthology}
\bibliographystyle{acl_natbib}

\appendix

% \clearpage
%======================================================================
\section{Appendix}
\label{Sec:appendix}
%======================================================================

%=================================
\subsection{Experimental Setups}
\label{sec:expSetup}
%=================================
% \balance

\paragraph{Dependencies.} We build and run our system with Pytorch 1.9.0 and AllenNLP 0.9.0 framework. We collect all verbs from the sentences in all datasets through spaCy\footnote{https://spacy.io/} POS tagger.
In addition, we obtain the constituency annotations through Stanford CoreNLP\footnote{https://stanfordnlp.github.io/CoreNLP/} and the dependency annotations through spaCy. We have total 27 types of constituency labels and 45 types of dependency labels.

\paragraph{Verb-tuple Alignment.}
We assume that every tuple has a verb in its relation (see Section~\ref{sec:model}). However, this assumption does not mean that each verb in a sentence can lead to one tuple. 
If a sentence contains multiple verbs identified by the POS tagger, we create multiple training instances. In each training instance, one verb is considered as the relation indicator, \ie its $W_{verb}$ initialized to 1, while $W_{verb}$ for all other verbs in the sentence are initialized to 0. The corresponding tuple taking this verb as relation is the gold label for this training instance. If this verb does not lead to a tuple, we set the label for all words in the sentence to be \say{O}, \ie no tuple extracted for this verb. As such, the model is able to learn which verb leads to a tuple extraction. During testing, multiple test instances are created if a sentence contains multiple verbs; one verb serves as a relation indicator in each test instance. No tuple is extracted if all predictions of this test instance are \say{O}.

\paragraph{Parameters.}
The hidden dimension $d_h$ for BERT representation $h_i^{bert}$, Dep-GCN graph representation $h_i^{dep}$, and Const-GCN graph representation $h_i^{con}$ is 768. We use single-layer GCNs for both constituency and dependency graphs. The hidden dimension $d_l$ for Dep-Encoder type embedding $l_i^{dep}$ and Const-Encoder path embedding $l_i^{con}$ is 400. Hyper-parameters $\alpha, \beta,  \gamma$ are set to 0.024, 0.012, and 0.012, respectively. Hyper-parameters selection is based on grid searching. The experiments are conducted with Tesla V100 32GB GPU and Intel$^\circledR$  Xeon$^\circledR$ Gold 6148 2.40 GHz CPU.

%=================================
\subsection{Re-implementation Details}
\label{sec:reimplementation}
%=================================
Note that all baselines are implemented to extract $n$-ary tuples on LSOIE-wiki and LSOIE-sci datasets, and binary tuples on CaRB dataset. `CopyAttention', `IMoJIE', and `CIGL-OIE + IGL-CA' are binary OpenIE systems and cannot be tested naturally on LSOIE-wiki and LSOIE-sci datasets. We re-implement their models to cater $n$-ary tuple extraction based on the code repositories.\footnote{The source code of `CopyAttention' and `IMoJIE' can be found in https://github.com/dair-iitd/imojie. The source code of `CIGL-OIE + IGL-CA' can be found in https://github.com/dair-iitd/openie6}
In the evaluation, we evaluate `CopyAttention', `IMoJIE', and `CIGL-OIE + IGL-CA' on LSOIE-wiki and LSOIE-sci datasets through our $n$-ary re-implementations.

\end{document}